\documentclass[runningheads]{llncs}
\usepackage{graphicx}
\usepackage{amsmath,amssymb} 
\usepackage{color}
\usepackage[width=122mm,left=12mm,paperwidth=146mm,height=193mm,top=12mm,paperheight=217mm]{geometry}
\usepackage{enumitem}

\newcommand{\hla}[1]{{\textcolor{black}{#1}}}

\iftrue
\newcommand{\todo}[1]{{\textcolor{red}{[[TODO: #1]]}}}
\newcommand{\outline}[1]{{\textcolor{blue}{[[#1]]}}}
\newcommand{\commenttext}[1]{\textcolor{red}{[[#1]]}}
\newcommand{\commentfoot}[1]{\footnote{\textcolor{red}{#1}}}
\newcommand{\commentselfoot}[2]{{\textcolor{blue}{#1}}\comment{#2}}
\newcommand{\commentselrep}[2] {{\textcolor{blue}{#1}} {\textcolor{green}{[[\textit{#2}]]}}}
\else
\newcommand{\hla}[1]{{#1}}
\newcommand{\todo}[1]{}
\newcommand{\outline}[1]{}
\newcommand{\commenttext}[1]{}
\newcommand{\commentfoot}[1]{}
\newcommand{\commentselfoot}[2]{}
\newcommand{\commentselrep}[2]{}
\fi

\iftrue

\newcommand{\cutsectionup}{}
\newcommand{\cutsectiondown}{}

\newcommand{\cutsubsectionup}{}
\newcommand{\cutsubsectiondown}{}

\newcommand{\cuttableup}{}
\newcommand{\cuttabledown}{}

\else

\newcommand{\cutsectionup}{\vspace*{-0.11in}}
\newcommand{\cutsectiondown}{\vspace*{-0.12in}}

\newcommand{\cutsubsectionup}{\vspace*{-0.11in}} 
\newcommand{\cutsubsectiondown}{\vspace*{-0.08in}} 

\newcommand{\cuttableup}{\vspace*{-0.15in}}
\newcommand{\cuttabledown}{\vspace*{-0.08in}}

\fi

\begin{document}
\pagestyle{headings}
\mainmatter

\title{Title Generation for User Generated Videos} 

\titlerunning{Title Generation for User Generated Videos}

\authorrunning{Kuo-Hao Zeng, Tseng-Hung Chen,	Juan Carlos Niebles,	Min Sun}

\author{Kuo-Hao Zeng$^1$, Tseng-Hung Chen$^1$,	Juan Carlos Niebles$^2$,	Min Sun$^1$}


\institute{$^1$Departmant of Electrical Engineering, National Tsing Hua University\\
$^2$Department of Computer Science, Stanford University\\
$^1$\email{\{s103061614@m103,s104061544@m104,sunmin@ee\}.nthu.edu.tw\\
$^2$jniebles@cs.stanford.edu}
}

\maketitle

\begin{figure}[t]
    	\includegraphics[width=1\textwidth]{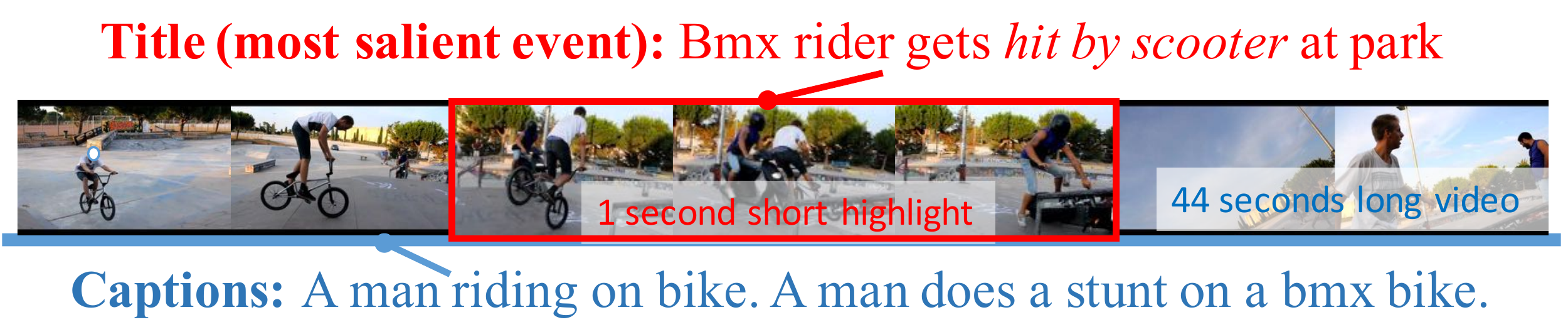}
    	\caption{\small Video title (top-red) v.s. video captions (bottom-blue) of a typical user generated video.
    	A video title describes the most salient event, which typically corresponds to a short highlight (1 sec in red box).
    	A caption describes a video as a whole (44 secs).
    	For a long video, there are many relevant captions, since many events have happened.
    	In this example, ``hit by scooter'' is a key phrase associated to the most salient event, while
        captions tend to be more generic to the overall contents of the sequence.
    	}\label{fig:teaser}
\end{figure}

\begin{abstract}
A great video title describes the most salient event compactly and captures the viewer's attention.
In contrast, video captioning tends to generate sentences that describe the video as a whole.
Although generating a video title automatically is a very useful task, it is much less addressed than video captioning.
We address video title generation for the first time by proposing two methods that extend state-of-the-art video captioners to this new task.
First, we make video captioners \emph{highlight sensitive} by priming them with a highlight detector.
Our framework allows for jointly training a model for title generation and video highlight localization.
Second, we induce high sentence diversity in video captioners, so that the generated titles are also diverse and catchy.
This means that a large number of sentences might be required to learn the sentence structure of titles.
Hence, we propose a novel \emph{sentence augmentation} method to train a captioner
with additional sentence-only examples that come without corresponding videos.
We collected a large-scale \emph{Video Titles in the Wild} (VTW) dataset of $18100$ automatically crawled user-generated videos and titles.
On VTW, our methods consistently improve title prediction accuracy,
and achieve the best performance in both automatic
and human evaluation.
\hla{Finally,
    our sentence augmentation method
    also outperforms the
    baselines on the M-VAD dataset.
}
\end{abstract}

\cutsectionup
\section{Introduction}
\cutsectiondown

Generating a natural language description of the visual contents of a video is one of the
holy grails in computer vision. Recently, thanks to breakthroughs in
deep learning~\cite{NIPS2012Alex} and Recurrent Neural Networks (RNN),
many attempts \cite{LSTMNAACL15,VideoAttRNN,SS15} have been made to jointly model videos and their corresponding sentence descriptions.
This task is often referred to as video captioning.
Here, we focus on a much more challenging task: \emph{video title generation}.
A great video title compactly describes the most salient event as well as catches people's attention
(e.g., ``bmx rider gets hit by scooter at park'' in Fig.~\ref{fig:teaser}-Top).
In contrast, video captioning generates a sentence to describe a video as a whole (e.g., ``a man riding on bike'' in Fig.~\ref{fig:teaser}-Bottom).
Video captioning has many potential applications such as helping the visually impaired to interpret the world.
We believe that video title generation can further enable Artificial Intelligence systems to communicate more naturally
by describing the most salient event in a long and continuous visual observation.

Video title generation poses two main challenges for existing video captioning methods \cite{VideoAttRNN,SS15}.
First of all, most video captioning methods assume that every video is trimmed into a 10-25 seconds short clip in both training and testing.
However, the majority of videos on the web are untrimmed, such as User-Generated Videos (UGVs) which are typically 1-2 minutes long.
The task of video title generation is to learn from untrimmed video and title pairs
to generate a title for an unseen untrimmed video.
In training, the first challenge is to temporally align a title to the most salient event, i.\@e.\@ the video highlight (red box in Fig.~\ref{fig:teaser}) in the untrimmed video.
Most video captioning methods, which ignore this challenge, are likely to learn an imprecise association between words and frequently observed visual evidence in the whole video.
Yao et al. \cite{VideoAttRNN} recently propose a novel soft-attention mechanism to softly select visual observation for each word.
However, we found that the learned per-word attention is prone to imprecise associations given untrimmed videos.
Hence, it is important to make video title generators ``highlight sensitive''.
As a second challenge, title sentences are extremely diverse (e.g., each word appears in only $2$ sentences on average in our dataset).
\hla{Note that the two latest movie description datasets \cite{TorabiPLC15,rohrbach2015dataset} also share the same challenge of diverse sentences.
On these datasets, state-of-the-art methods \cite{VideoAttRNN,SS15} have reported fairly low performance.}
Hence, it is important to ``increase the number of sentences'' for training a more reliable language model. We propose two generally applicable methods to address these challenges.

\noindent\textbf{Highlight sensitive captioner.}
We combine a highlight detector with video captioners~\cite{VideoAttRNN,SS15} to train models that can jointly generate titles and locate highlights.
The highlights annotated in training can be used to further improve the highlight detector.
As a result, our ``highlight sensitive'' captioner learns to generate title sentences specifically describing the highlight moment in a video.

\noindent\textbf{Sentence augmentation.}
To encourage the generation of more diverse titles, we augment the training set with sentence-only examples that do not come with corresponding videos.
Our intuition is to learn a better language model from additional sentences.
In order to allow state-of-the-art video captioners to train with additional sentence-only examples,
we introduce the idea of ``dummy video observation''.
In short, we associate all augmented sentences to the same dummy video observation in training
so that the same training procedures in most state-of-the-art methods (e.g., \cite{VideoAttRNN,SS15})
can be used to train with additional augmented sentences.
This method enables any video captioner to be improved by
observing additional sentence-only examples, which are abundant on the web.

To facilitate the study of our task, we collected a challenging large-scale ``Video Title in the Wild'' (VTW) dataset\footnote{VTW dataset can be accessed at \url{http://aliensunmin.github.io/project/video-language/}} with the following properties:

\noindent \textbf{Highly open-domain.} Our dataset consists of  $18100$ automatically crawled UGVs as opposed to self-recorded single domain videos \cite{rohrbach13iccv}.

\noindent \textbf{Untrimmed videos.} Each video is on an average 1.5 minutes (45 seconds median duration) and contains a highlight event which makes this video interesting.
Note that our videos are almost 5-10 times longer than clips in \cite{TorabiPLC15}.
Our highlight sensitive captioner precisely addresses the unknown highlight challenge.

\noindent \textbf{Diverse sentences.} Each video in our dataset is associated with one title sentence. The vocabulary is very diverse, since on average each word only appears in $2$ sentences in VTW, compared to $5.3$ sentences in \cite{chen:acl11}.
Our sentence augmentation method directly addresses the diverse sentences challenge.

\noindent \textbf{Description.} Besides titles, our dataset also provides accompanying description sentences with more detailed information about each video. These sentences differ from the multiple sentences in \cite{chen:acl11}, since our description may refer to non-visual information of the video. We show in our experiments that they can be treated as augmented sentences to improve video title generation performance.

We address video title generation with the following contributions.
(1) We propose a novel highlight sensitive method to adapt two state-of-the-art video captioners \cite{VideoAttRNN,SS15} to video title generation. Our method significantly outperforms \cite{VideoAttRNN,SS15} in METEOR and CIDEr.
(2) Our highlight sensitive method improves highlight detection performance from $54.2\%$ to $58.3\%$ mAP.
(3) We propose a novel sentence augmentation method to train state-of-the-art video captioners with additional sentence-only examples. This method significantly outperforms \cite{VideoAttRNN,SS15} in METEOR and CIDEr.
(4) We show that sentence augmentation can be applied on another video captioning dataset (M-VAD~\cite{TorabiPLC15}) to further improve the captioning performance in METEOR.
(5) By combining both methods, we achieve the best video title generation performance of $6.2\%$ in METEOR and $25.4\%$ in CIDEr.
(6) Finally, we collected one of the first large-scale ``Video Title in the Wild" (VTW) dataset to benchmark the video title generation task. The dataset will be released for research usage.

\cutsectionup
\section{Related Work}
\cutsectiondown

\noindent \textbf{Video Captioning.}
Early work on video captioning \cite{rohrbach13iccv,CorsoCVPR13,GuadarramaICCV13,krishnamoorthy:aaai13,thomason:coling14,VISO,Kojima2002} typically perform a two-stage procedure.
In the first stage, classifiers are used to detect objects, actions, and scenes. In the second stage, a model combining visual confidences with a language model is used to estimate the most likely combination of subject, verb, object, and scene. Then, a sentence is generated according to a predefined template.
These methods require a few manual engineered components such as the content to be classified and the template.
Hence, the generated sentences are often not as diverse as sentences used in natural human description.

Recently, image captioning methods \cite{lrcn2014,showtell2015,xu2015show,mao2015learning,mao2015generation,johnson2015densecap} begin to adopt the Convolutional Neural Networks (CNN) and Recurrent Neural Networks (RNN) approaches. They learn models directly from a large number of image and sentence pairs. The CNN replaces the predefined features to generate a powerful
distributed visual representation. The RNN takes the CNN features as input and learns to decode it into a sentence.
These are combined into a large network that can be jointly trained to directly map an image to a sentence.

Similarly, recent video captioning methods adopt a similar approach.
Venugopalan et al.~\cite{LSTMNAACL15} map a video into a fix dimension feature by average-pooling CNN features of many frames
and then use a RNN to generate a sentence. However, this method discards the temporal information of the video.
Rohrbach et al.~\cite{GCPR15} propose to combine different RNN architectures with multiple CNN classifiers for classifying verbs (actions), objects, and places.
Lisa Anne Hendricks et al.~\cite{hendricks2015deep} propose to utilize unpaired data for training to generate image captions and video descriptions.
To capture temporal information in a video, Venugopalan et al. \cite{SS15} propose to use RNN to encode a sequence of CNN features extracted from frames following the temporal order. This direct video-encoding and sentence-decoding approach outperforms \cite{LSTMNAACL15} significantly.
Concurrently, Yao et al.~\cite{VideoAttRNN} proposes to model the temporal structure of visual features in two ways.
First, it designs a 3D CNN based on dense trajectory-like features \cite{wang:2011} to capture local temporal structure.
Then, it incorporates a soft-attention mechanism to select temporal-specific video observations for generating each word.
Our proposed highlight sensitive method can be considered as a hard-attention mechanism to select a video segment (i.e., a highlight) for generating the sentence.
In our experiments, we find that our highlight sensitive method further improves \cite{VideoAttRNN}.
Instead of RNN for encoding or decoding, Xu et al.~\cite{xu2015jointly} propose to embed both video and sentence to a joint space.
{Most recently,
Pan et al.~\cite{PanMYLR15} further propose a novel framework to jointly perform visual-semantic embedding and learn a RNN model for video captioning. Pan et al.~\cite{PanXYWZ15} propose a novel Hierarchical RNN to exploit video temporal structure in a longer range.
Yu et al.~\cite{YuWHYX15} propose a novel hierarchical framework containing a sentence generator and a paragraph generator.
Despite many new advances in video captioning, video title generation has not been well studied.}

\noindent \textbf{Video Highlight Detection.}
Most early highlight detection works focus on broadcasting sport videos~\cite{LiuHLACCV96,AceroACMM00,ReyACMM01,TianICME04,HuangICME05,SenguptaICME06,AIEEEM05,googleHL11}.
Recently, a few methods have been proposed to detect highlights in generic personal videos.
Sun et al.~\cite{MsunECCVSM14} automatically harvest user preference to learn a model for identifying highlights in each domain.
Instead of generating a video title, Song et al.~\cite{song2015tvsum} utilize video titles to summarize each video.
The method requires additional images to be retrieved by title search for learning visual concepts.
There are also a few fully unsupervised approaches.
Zhao and Xing~\cite{zhao2014quasi} propose a quasi-real time method to generate short summaries.
Yang et al.~\cite{yang2015unsupervised} propose a recurrent auto-encoder to extract video highlights.
Our video title generation method is one of the first to combine explicit highlight detection (not soft-attention) with sentence generation.

\noindent \textbf{Video Captioning Datasets.}
A number of video captioning datasets~\cite{TorabiPLC15,rohrbach2015dataset,rohrbach13iccv,chen:acl11,CorsoCVPR13,rohrbach14gcpr,xu2016msr} have been introduced.
Chen and Dolan~\cite{chen:acl11} collect one of the first multiple-sentence video description datasets with 1967 YouTube videos.
The duration of each clip is between 10 and 25 seconds, typically depicting a single activity or a short sequence.
It requires significant human effort to build this dataset, since all $70028$ sentences are labeled by crowdsourced annotators.
On the other hand, we collect our dataset with a large number of video and sentence pairs fully automatically.
Rohrbach et al.~\cite{rohrbach2015dataset} collect a movie dataset with $54076$ sentences from audio transcripts and video snippets in 72 HD movies.
It also takes significant human effort to build this dataset, since each sentence is manually aligned to the movie.
Torabi et al.~\cite{TorabiPLC15} collect a movie dataset with $55904$ sentences from audio transcripts and video snippets in 96 HD movies.
They introduce an automatic Descriptive Video Service (DVS) segmentation and alignment method for movies. Hence, similar to our automatically collected dataset, they can scale up the collection of a DVS-derived dataset with minimal human intervention.
Jun Xu et al.~\cite{xu2016msr} collect a large video description dataset by 257 popular queries from a commercial video search engine, with 118 videos for each query.
We compare the sentences in our dataset with two movie description datasets in Sec.~\ref{sec.DataComp} and
find that our vocabularies are fairly different \hla{(see \cite{VTGtech})}. In this sense, our dataset is complementary to theirs.
However, both datasets are not suitable for evaluating video title generation, since they consist of short clips with 6-10 seconds and selecting the most salient event in the video is not critical.

\cutsectionup
\section{Video Title Generation}\label{sec.3}
\cutsectiondown

Our goal is to automatically generate a title sentence for a video, where the title should compactly describe the most salient event in the video.
This task is similar to video captioning, since both tasks generate a sentence given a video.
However, most video captioning methods focus on generating \emph{a relevant sentence} given a 6-10 seconds short clip.
In contrast, video title generation aims to produce a title sentence describing the most salient event given a typical 1 minute user-generated video (UGV).
Hence, video title generation is an important extension of generic video captioning
to understand a large number of UGVs on the web.

To study video title generation, we have collected a new ``Video Titles in the Wild" (VTW) dataset that consists of UGVs.
We first introduce the dataset and discuss its unique properties and the challenges for video title generation.
Then, our proposed methods will be introduced in Sec.~\ref{sec.m}.

\begin{figure}[!t]
\cuttableup
    	\includegraphics[width=1\textwidth]{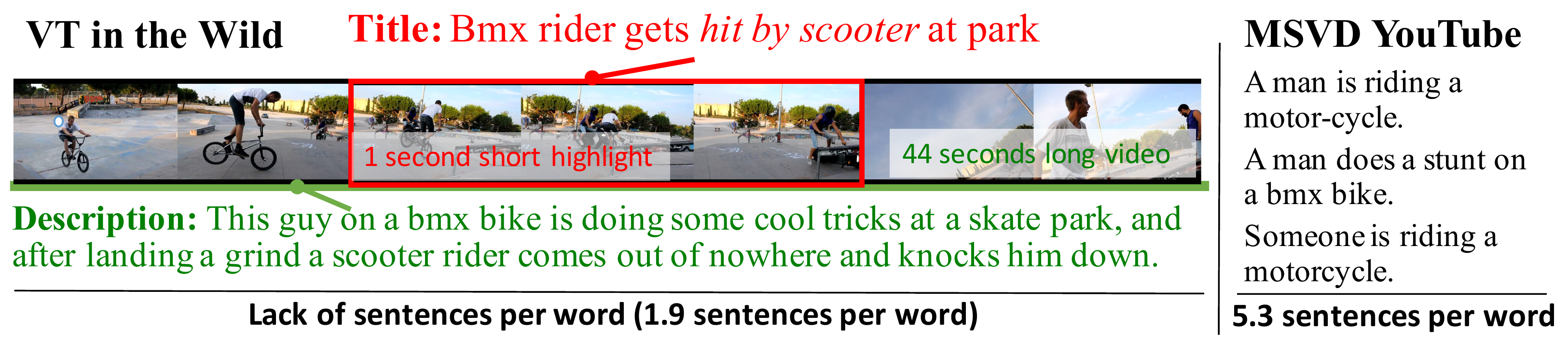}
    	\caption{\small Dataset comparison. Left-panel: VTW.  Right-panel: the MSVD~\cite{chen:acl11}.
    	}\label{fig:data}
\cuttabledown
\end{figure}

\cutsubsectionup
\subsection{Collection of Curated UGVs}\label{sec.UGV}
\cutsubsectiondown
Everyday, a vast amount of UGVs are uploaded to video sharing websites.
To facilitate web surfers to view the interesting ones, many online communities curate a set of interesting UGVs.
We program a web crawler to harvest UGVs from these communities.
For this paper, we have collected $18100$ open-domain videos
with 1.5 minutes duration on average (45 seconds median duration).
We also crawl the following curated meta information about each video (see Fig.~\ref{fig:data}):
\emph{Title:} a single and concise sentence produced by an editor, which we use as ground truth for training and testing;
\emph{Description:} 1-3 longer sentences which are different from titles, as they may not be relevant to the salient event, or may not be relevant to the visual contents;
\emph{Others:} tags, places, dates and category.

This data is automatically collected from well established online communities that post 10-20 new videos per day. We do not conduct any further curation of the videos or sentences so the data can be considered ``in the wild''.

\noindent \textbf{Unknown Highlight in UGVs.}\label{sec.DHL}
We now describe how title generation is related to highlight in UGVs.
These UGVs are on an average $1.5$ minutes long which is 5-10 times longer than
clips in video captioning datasets \cite{TorabiPLC15,rohrbach2015dataset}.
Intuitively, the title should be describing a segment of the video corresponding to the highlight (i.e., the salient event).
To confirm this intuition, we manually label title-specific highlights (i.e., compact video segments well described by the titles) in a subset of videos.
We found that the median highlight duration is about $\hla{3.3}$ seconds.
Moreover, the non-highlight part of the video might not be precisely described by the title.
In our dataset, the temporal location and extent of the highlight in most videos are unknown.
This creates a challenge for a standard video captioner to learn the correct association between words in titles and
video observations.
In Sec.~\ref{sec.THL}, we propose a novel highlight-sensitive method to jointly locate highlights and generate titles for addressing this challenge.

\cutsubsectionup
\subsection{Dataset Comparison}\label{sec.DataComp}
\cutsubsectiondown

\begin{table*}[!t]
\cuttableup
\hspace{-0mm}
   \centering
   \caption{Dataset Comparison. Our data is from a large-scale open-domain video repository and our total duration is 2.5 times longer than \cite{TorabiPLC15}.
    V. stands for video, and (V) denotes videos of a few minutes long, whereas clips are typically a few seconds long. Desc. stands for description. AMT stands for Amazon Mechanical Turk. DVS stands for Descriptive Video Service.}
    \begin{tabular}{|c|c|c|c|c|c|}
    \hline
    \hspace{-0mm}Dataset & V. Source & \#Clips & Duration (H) & Desc. Source & \#Sentences \\ \hline
    \hspace{-0mm}YouCook \cite{CorsoCVPR13} & Cooking & 88 (V) & 2.3& AMT &2,668\\ \hline
    \hspace{-0mm}TACoS \cite{rohrbach13iccv} & Cooking & 7,206 & 15.9& AMT & 18,227\\ \hline
    \hspace{-0mm}TACoS-M \cite{rohrbach14gcpr} & Cooking & 14,105 & 27.1 & AMT & 52,593\\ \hline\hline
	\hspace{-0mm}MPII-MD. \cite{rohrbach2015dataset} & Movie & 54,076 & 56.5 & Script + DVS & 54,076\\ \hline
	\hspace{-0mm}(DVS part) \cite{rohrbach2015dataset} & Movie & 30,680 & 34.7& DVS & 30,680\\ \hline
	\hspace{-0mm}M-VAD \cite{TorabiPLC15} & Movie & 48,986 & 84.9 & DVS & 55,904\\ \hline\hline
	\hspace{-0mm}MSVD \cite{chen:acl11} & YouTube & 1,970 & $\sim$9.6\footnote{Each video is 10 to 25 seconds. We assume each video is $(10+25)/2=17.5$ seconds.}& AMT & 70,028\\ \hline
	\hspace{-0mm}VTW-title & YouTube &18,100 (V) & 213.2 & Editor & 18,100 \\ \hline
   \hspace{-0mm}VTW-full & YouTube &18,100 (V) & 213.2 & Owner/Editor & 44,603\\ \hline
    \end{tabular}
    \label{table.comp}
\cuttabledown
\end{table*}

Our VTW dataset is a challenging large-scale video captioning dataset, as summarized in Table \ref{table.comp}.
The VTW dataset has the longest duration ({213.2} hours) and each of our videos is about $10$ times longer than each clip in \cite{TorabiPLC15,rohrbach2015dataset}.
The table also shows that only movie description datasets \cite{TorabiPLC15,rohrbach2015dataset} and VTW are:
(1) at the scale of more than $10K$ open-domain videos, and (2) consisting of sophisticated sentences produced by editors instead of simple sentences produced by Turkers.

\begin{table}[!b]
\cuttableup
   \centering
   \caption{Text Statistics. The first two columns are the number of sentences and non-stemmed vocabulary size, respectively. The third column is the average number of sentences per word. The last four columns are nouns, verbs, adjectives, and adverbs in order, where A;B denotes A as number and B as ratio. We compute the ratio with respect to the number of nouns. Voca. stands for vocabulary. Sent. stands for sentences. W. stands for words. Our full dataset has vocabulary with a similar size compared to two recent large-scale video description datasets.}
   \small
    \begin{tabular}{|l|c|c|c|c|c|c|c|c|}
    \hline
    \hspace{-0mm}& \#Sent. & Voca. & \#Sent./W.  & \#Nouns & \#Verbs & \#Adjective & \#Adverb \\ \hline
	\hspace{-0mm}MPII-MD \cite{rohrbach2015dataset} & 54076 &  20650& 2.6 &  11397;1 & 6100;0.54&  3952;0.35& 1162;0.1 \\ \hline
	\hspace{-0mm}M-VAD \cite{TorabiPLC15} & 55904 & 18310 & 3.0& 10992;1 & 4945;0.45  &3649;0.33  & 870;0.08 \\ \hline
	\hspace{-0mm}VTW-title & 18100 & 8874& 2.0 & 5850;1  & 2187;0.37 & 1187;0.2  &  224;0.04 \\ \hline
	\hspace{-0mm}VTW-full & 44603 & 23059& 1.9 &13606;1  & 6223;0.46 &3967;0.29  &  846;0.06 \\ \hline
    \end{tabular}
    \label{table.POS}
\cuttabledown
\end{table}

\noindent \textbf{Sentence diversity.}
Intuitively, a set of diverse sentences should have a large vocabulary.
Hence, we use the ratio of the number of sentences to the size of vocabulary as a measure of sentence diversity.
We found that the MSVD dataset has on an average $5.3$ sentences per word,
whereas both movie description datasets have less than or equal to $3$ sentences per-word and VTW has about $2$ sentences per word (Table.~\ref{table.POS}).
Therefore, sentences in VTW are twice more diverse than in the MSVD dataset and slightly more diverse than in the movie description datasets.
This implies that we need more sentences for learning, even though these datasets are already the largest datasets.
In Sec.~\ref{sec.SA}, we propose a novel ``sentence augmentation" method to mitigate this issue.

\noindent \textbf{Complementary vocabulary.}
Although the distribution of nouns, verbs, adjectives, and adverbs in all three datasets are similar (see Table \ref{table.POS}),
the common words are different in these two types of datasets, since VTW consists of UGVs and \cite{rohrbach2015dataset,TorabiPLC15} consists of movie clips.
We visualize the top few nouns and verbs in VTW, MPII-MD.~\cite{rohrbach2015dataset}, and M-VAD~\cite{TorabiPLC15} \hla{in the technical report~\cite{VTGtech}}. We believe our dataset is complementary to the movie description datasets for future study of both video captioning and title generation.

\cutsectionup
\section{From Caption to Title}\label{sec.m}
\begin{figure}[!t]
\cuttableup
    	\includegraphics[width=1\textwidth]{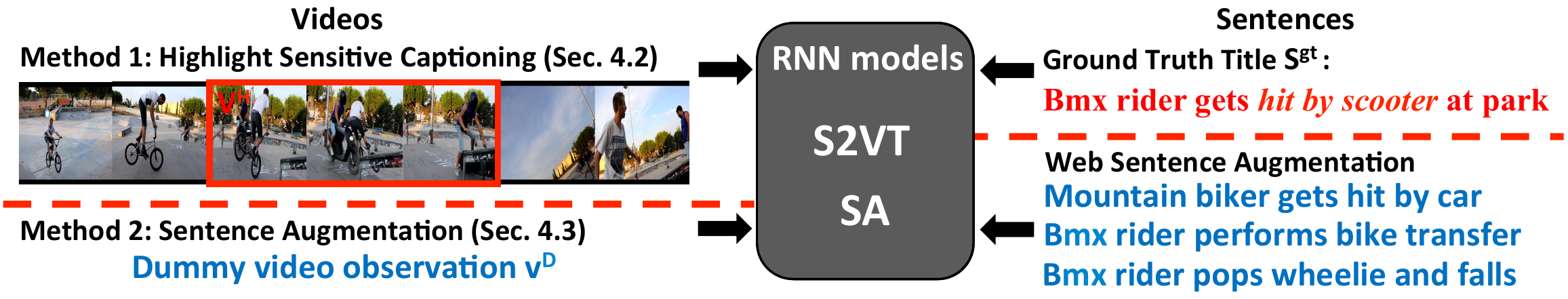}
    	\caption{\small An overview of our proposed methods: (top-row) highlight sensitive captioning (Sec.~\ref{sec.THL}) and (bottom-row) sentence augmentation
    	(Sec.~\ref{sec.SA}).
    	}\label{fig:system}
\cuttabledown
\end{figure}

\cutsectiondown
Both video title generation and captioning models
learn from many video $V$ and sentence $S$ pairs,
where $V$ contains a sequence of observations $(v_1,\dots,v_k,\dots,v_n)$ and $S$ a sequence of words $(s_1,\dots,s_i,\dots,s_m)$.
In this section, we build from the video captioning task and introduce two generally applicable methods (see Fig.~\ref{fig:system}) to handle the challenges for video title generation.

\cutsubsectionup
\subsection{Video Captioning}
\cutsubsectiondown
Video captioning can be formulated as the following optimization problem,
\small
\begin{eqnarray}
{S}^*(V;\theta)=\arg\max_{S} p(S|{V};\theta)~,
\end{eqnarray}
\normalsize
where $S^*(V;\theta)$ is the predicted sentence, $\theta$ is the learned model parameters,
and $p(S|V;\theta)$ is the conditional probability of sentence $S$ given a video sequence $V$.
According to the probability chain rule, the full sentence conditional probability $p(S|V;\theta)$ equals to the multiplication of each word conditional probability:
\small
\begin{eqnarray}
p(S|{V};\theta)=\prod_{i=1}^{m}p(s_i|S_{1:(i-1)},V)~,
\end{eqnarray}
\normalsize
where $s_i$ is the $i^{th}$ word, $S_{1:(i-1)}$ is the partial sentence from the first word to the $i-1^{th}$ word.
Note that the $i^{th}$ word depends on all the previously generated words $S_{1:(i-1)}$ and the video $V$.
Most state-of-the-art methods utilize Recurrent Neural Networks with Long Short Term Memory (LSTM) cells~\cite{LSTM} to
model the long-term dependency in this single word conditional probability.
We use two state-of-the-art methods as examples,
\small
\begin{itemize}[leftmargin=*]
\item{Sequence to Sequence - Video to Text (S2VT) \cite{SS15}.}  The method proposed to use RNN to encode both the video sequence $V=(v_1,\dots,v_k,\dots,v_n)$
and partial sentences $S_{1:(i-1)}=(s_1,\dots,s_{i-1})$ into a learned hidden representation $h_{n+i-1}$ so that the single word conditional probability
becomes $p(s_i|h_{n+i-1},s_{i-1})$.
\item{Soft-Attention (SA) \cite{VideoAttRNN}.} The model proposed to use RNN to encode the partial sentences $S_{1:(i-1)}=(s_1,\dots,s_{i-1})$ into a learned hidden representation $h_{i-1}$ and apply per-word soft-attention mechanism to obtain weighted average of all video observation $\varphi(V)=\sum_{i}^n \alpha_iv_i$, where $\sum_{i}^n \alpha_i = 1$. The single word conditional probability becomes $p(s_i|h_{i-1},s_{i-1},\varphi(V))$.
\end{itemize}
\normalsize
Despite their differences, they essentially model two relations:
\small
\begin{itemize}[leftmargin=*]
\item{Word and video ($s_i|V$).} This relation is critical for associating words to video observation. However, this relation alone is only sufficient for video tagging, but not video captioning.
\item{Words sequence ($s_i|S_{1:(i-1)}$).} Modeling this relation is the essence of language modeling. However, this relation alone is only sufficient for sentence generation (i.e., captioning), but not video captioning.
\end{itemize}
\normalsize
An ideal video captioning method should model both types of relations equally well.
In particular, our video title generation task creates additional challenges on modeling these relations: (1) unknown highlight, (2) diverse sentences.
We now present our novel and generally applicable methods for improving the modeling of these two relations for video title generation.

\cutsubsectionup
\subsection{Highlight Sensitive Captioning}\label{sec.THL}
\cutsubsectiondown
As we mentioned in Sec.~\ref{sec.DHL},
UGVs are on an average 1.5 minutes with many parts
not precisely described by the title sentence.
Hence, it is very challenging to learn the right $s|V$ relation given many irrelevant video observations in $V$.
Intuitively, there should exist a video highlight $V^H \subset{V}$ which is the most relevant to the ground truth title sentence $S^{gt}$ (see Fig.~\ref{fig:system}-Top).
We propose to train a highlight sensitive captioner by solving the following optimization problem,
\small
\begin{eqnarray}
\arg\min_{\theta ,\{V_j^H\}_j}\sum_j \mathcal{L}(S_j^{gt},S_j^*(V_j^H;\theta))\textit{   ;   }
\mathcal{L}(S^{gt},S^*(V;\theta)) =\sum_{i} L(s_i^{gt},s_i^*(V;\theta))~,
\end{eqnarray}
\normalsize
where $j$ is the video index (omitted for conciseness in many cases), $S^*(V;\theta)$ is the predicted sentence given the video $V$ and model parameter $\theta$, $i$ is the word index, $s^{gt}_i$ is the ground truth $i^{th}$ word, $s^*_i$ is the predicted $i^{th}$ word, and $L$ is the cross-entropy loss.
This is a hard optimization problem, since jointly optimizing the continuous variable $\theta$ and discrete variables $\{V_j^H\}_j$ is NP-hard.
However, when video highlights $\{V_j^H\}_j$ are fixed, the optimization problem is the original video captioning problem.

\noindent \textbf{Training procedure.}
We propose to iteratively solve for $\theta$ and $V^H$.
When $V_H$ is fixed, we use stochastic gradient descent to solve for $\theta$.
Next, when $\theta$ is fixed, we use the loss $\mathcal{L}(.)$ to find the best $V^{H*}$ by solving,
\small
\begin{eqnarray}
V^{H*} = \arg\min_{V^H \in V} \mathcal{L}(S^{gt},S^*(V;\theta))~.\label{eq.H}
\end{eqnarray}
\normalsize
The training loss typically converged within a few iterations, since $p(.)$ is a deep model with high-capacity.
This implies that our iterative training procedure needs to start with a good initialization.
We propose to train a highlight detector on a small set of training data with ground truth highlight labels.
Then, use the detector to automatically obtain the initial video highlight $V^H$ on the whole training set to start the iterative training procedure.

At each iteration, the updated highlight $V^H$ can be used to (1) retrain the highlight detector using the full training set, and (2) update the video captioning model.
As a result, our ``highlight sensitive'' captioner learns to generate sentences specifically describing the highlight moment in a video.
We found that the refined highlight detector achieves a better performance.

\cutsubsectionup
\subsection{Sentence Augmentation}\label{sec.SA}
\cutsubsectiondown
As mentioned above, we are facing the lack of sentences issue due to the diverse sentence property.
We argue that the ability to jointly train the captioner with sentence-only examples (with no corresponding videos) and video-sentence pairs
is a critical strategy to increase the robustness of the language model.
However, most state-of-the-art captioners~\cite{SS15,VideoAttRNN} are strictly trained with video-sentence pairs only.
This prevents video captioning to benefit from other sentence-only information on the web.
Moreover, we confirm in experiment that a video-description pairs training procedure does not consistently improve performance.
Hence, we propose a novel and generally applicable method to train a RNN model with both video-sentence pairs and sentence-only examples,
where sentence-only examples are either the description sentences or additional sentences on the web.
The idea of our technique is straight forward: let's associate a dummy video observation $v^D$ to a sentence-only example (see Fig.~\ref{fig:system}-Bottom).

\noindent \textbf{Dummy video observation.}
We design the dummy video observation $v^D$ for SA~\cite{VideoAttRNN} and S2VT~\cite{SS15}, separately, by considering their model structures.

In SA, all video observations are weighted summed into a single observation $\varphi(V)=\sum_{i}^n \alpha_iv_i$, where $\sum_{i}^n \alpha_i = 1$.
The video observation $\varphi(V)$ is, then, embedded to $A\varphi(V)$ in the LSTM cell.
For the augmented sentences with no corresponding video observations, we design $v_i=v^D$ as an all zeros vector except a single $1$ at the first entry and let it be a constant observation across time.
This implies that $A\varphi(\{v^D\})=Av^D=a^1$, where $A=[a^1,\dots]$.
Intuitively, $a^1$ can be considered as a trainable bias vector to handle additional sentence-only examples.
As a concrete example, the memory cell in SA is updated as below,
\small
\begin{eqnarray}
c_t=\textrm{tanh}(W_cE[y_{t-1}]+U_c h_{t-1}+A_c\varphi(\{v^D\})+b_c)~,
\end{eqnarray}
\normalsize
where $c_t$ is the new memory content, $E[y_{t-1}]$ is the previous word, $h_{t-1}$ is the previous hidden representation, 
$W_c,U_c,A_c$ are trainable embedding matrices, and $b_c$ is the original trainable bias vector.
Now $A_c\varphi(\{v^D\})=a_c^1$ can be considered as another trainable bias vector to handle the dummy video observations.

In S2VT, all video observations are sequentially encoded by RNN as well. However, if we design the $v^D$ as an all zeros vector except a single $1$ at the first entry,
the encoded representation $h_n$ at the end of the video sequence will be a function of all model parameters: $W_{x*}$, $W_{h*}$ and $b_*$.
Hence, we simply design $v^D$ as an all zeros vector so that $h_n$ will be a function of only $W_{h*}$ and $b_*$.
Intuitively, this simplifies the parameters that handle additional sentence-only examples with dummy video observations.
In our experiments, we find that the all zeros vector achieves a better accuracy for S2VT \hla{(see ~\cite{VTGtech} for details)}.

 \cutsectionup
\section{Experiments}\label{sec.exp}
\cutsectiondown

We first describe general details of our experimental settings and implementation.
Then, we define variants of our methods and compare performance on VTW and \hla{M-VAD~\cite{TorabiPLC15}}.

\noindent \textbf{Benchmark Dataset.}
We randomly split our dataset into $80\%$ training, $10\%$ validation, and $10\%$ testing as the same proportion in the M-VAD~\cite{TorabiPLC15}.
In this paper, we mainly use title sentences.
This means we have $14100$ video-sentence pairs for training, $2000$ pairs for validation, and $2000$ pairs for testing.
Our dataset is extremely challenging: among $2980$ unique words in testing, there are $488$ words ($16.4\%$) which have not appeared in training, $323$ words ($10.8\%$) which have only appeared once in training.
\hla{We refer these numbers as ``Testing-Word-Count-in-Training" (TWCinT) statistics and show these statistics in the technical report~\cite{VTGtech}.}
We also manually labeled the highlight moments in $2000$ training ($14.2\%$ of total training) and $2000$ testing ($100\%$ of total testing) videos.
These labels in the training set are only used as supervision to train the initial highlight detector.
These labels in the testing set are only used as ground truth for evaluating highlight detection accuracy.

\noindent \textbf{Features.}
Similar to existing video captioning methods, we utilize both appearance and local motion features:
we extract VGG~\cite{SimonyanICLR15} features for each frame, and
C3D~\cite{C3D15} features for 16 consecutive frames.
For S2VT~\cite{SS15} and SA~\cite{VideoAttRNN}, we embed both features to a lower 500 and 1024 dimension space, respectively,
according to their original papers.
Next, we define the video observation.

\noindent \textbf{Video observation.}
We divide a video into maximum 45-50 clips due to GPU memory limit,
and average-pool features within each clip.

\noindent \textbf{Highlight Detector.}
We train a bidirectional RNN highlight detector (details in \cite{VTGtech}) on $2000$ training videos to predict the highlightness of each clip of $100$ frames, since the median ground truth highlight duration is about $100$ frames.
This initial highlight detector achieves a {$54.2\%$} mean  Average Precision (mAP) on testing videos.
The trained detector selects eight consecutive highlight clips (800 consecutive frames) for each training video to train a captioner.
After a captioner is trained, it will select again eight consecutive clips as the highlight (see Eq.~\ref{eq.H}) to (1) retrain a highlight detector, and (2) a captioner.

\noindent \textbf{Sentence Augmentation.}
Given a large corpus, we retrieve additional sentences for sentence augmentation as follows.
We use each training sentence as a query and retrieve similar sentences in the corpus.
We use the mean of word2vec~\cite{mikolov2013efficient} feature of non-stop words in each sentence as the sentence-based feature.
Cosine similarity is used to measure sentence-wise similarity.
Among sentences with similarity above $0.75$, we sample a target number of sentences.
On VTW, we use $14100$ titles in training set to retrieve sentences from a corpus of YouTube video titles for augmentation.
In detail, we use YouTube API to download video titles in a few UGVs channels. There are 3549 unique sentences with a vocabulary of 3732 words.
\hla{On M-VAD, we retrieve {$23635$} sentences from MPII-MD~\cite{rohrbach2015dataset} for augmentation.}

\noindent \textbf{RNN training.}
In all experiments, we use \hla{$0.0001$} learning rate, \hla{$200$} maximum epochs, \hla{$10$} batch size, and stochastic gradient-based solver~\cite{KingmaB14} with its default parameters in TensorFlow~\cite{tensorflow2015-whitepaper} to train a model from scratch.
When finetuning a model, we train for another $50$ epochs.
Hence, HL requires additional $50 \times N$ epochs, where N is the the number of iteration, than Vanilla and HL-1. WebAug is trained with $200$ epochs but with a larger number of min-batches due to sentence augmentation.
All models are selected according to validation accuracy.

\noindent \textbf{Evaluation metric.}
We use the standard evaluation metric for the image captioning challenge~\cite{MSCOCO} including BLEU1 to BLEU4, METEOR, and CIDEr~\cite{Vedantam_2015_CVPR}.
METEOR is a metric replacing BLEU1 to BLEU4 into a single performance value, and it is designed to improve correlation with human judgments.
CIDEr is a new metric recently adopted for evaluating image captioning.
It considers the rareness of n-grams (computed by tf-idf), and gives higher value when a rare n-gram is predicted correctly.
Since typically a few important words make a title sentence stands out (e.g., hit by scooter in Fig.~\ref{fig:teaser}),
we also consider CIDEr as a good evaluation metric for video title generation.
\hla{Other than these automatic metrics, we also ask human judges to select the better video title out of a sentence generated by a state-of-the-art video captioner~\cite{SS15}
or a sentence  generated by our best method.}
\normalsize

\cutsubsectionup
\subsection{Baseline Methods}\label{sec.BM}
\cutsubsectiondown
We define variants of our methods for performance comparison.
\begin{itemize}[leftmargin=*]
\item{Vanilla represents our TensorFlow reimplementation of either S2VT~\cite{SS15} or SA~\cite{VideoAttRNN} (see technical report~\cite{VTGtech} for details). Note that these are two fairly strong baseline methods.}
\item{Vanilla-GT-HL denotes that ground truth highlight clips are used while evaluating the Vanilla model.}
\item{HL-1 denotes the initially trained highlight-sensitive captioner. Its comparison with Vanilla shows the effectiveness of highlight detection.}
\item{HL denotes the converged  highlight-sensitive captioner. At each iteration, we finetune the model from previous iteration.}
\item{Vanilla+Desc. treats descriptions as additional title sentences associated to their original videos in training. This is a risky assumption, since many descriptions describe the non-visual information of the videos.}
\item{Desc. Aug. uses descriptions as augmented sentences.}
\item{Web Aug. retrieves sentences from another corpus as augmented sentences.}
\item{HL+Web Aug. combines highlight sensitive captioning with sentence augmentation.
\hla{In detail, we take the trained Web Aug. model as the initial model. Then, we apply our HL method and finetune the model.}
}
\end{itemize}

\begin{table*}[!t]
\cuttableup
\centering
\caption{Video captioning performance of different variants of our methods (see Sec.~\ref{sec.BM}) on VTW dataset.
Our methods are applied on two state-of-the-art methods: S2VT~\cite{SS15} (Left-columns) and SA~\cite{VideoAttRNN} (Right-columns).
By combining highlight with sentence augmentation (HL+Web Aug.), we achieves the best accuracy consistently across all measures (highlight in bold-font).
MET. stands for METEOR. B@1 denotes BLEU at 1-gram.
Desc. stands for description. Aug. stands for sentence augmentation.}
\begin{tabular}{|c|c|c|c|c|c|c|c|c|c|c|c|c|}
\hline
     VTW                                                  & \multicolumn{6}{c|}{S2VT~\cite{SS15} (\%)}                 & \multicolumn{6}{c|}{SA~\cite{VideoAttRNN} (\%)}               \\  \hline
Variant & B@1 & B@2 & B@3 & B@4 & MET. & CIDEr & B@1 & B@2 & B@3 & B@4 & MET. & CIDEr \\  \hline
Vanilla                                                 &  9.3   &   3.7  &  1.9   &  1.2   &   5.2     &   18.6    &  9.2   & 4.1   & 2.2    &  1.4   &   4.5     &  18.5     \\  \hline
Vanilla-GT-HL                       &  10.2   &  4.3  &  2.1   &  1.2   &  5.1  &  19.8  &  9.4   & 4.3   & 2.3   &  1.5   &  4.7   &  19.8    \\  \hline
HL-1                                         &  10.8   &  4.5   &  2.3   &   1.4  &  6.1      &   23.0    &   11.6  &   \textbf{5.5}  &  \textbf{2.9}   &   1.7  &   5.6    &   24.3    \\  \hline
HL                                              &  11.4   &  4.9   &  2.5   &  \textbf{1.6}   &  \textbf{6.2}      &   {24.9}    &  11.6   &  5.3   &   \textbf{2.9}  &  1.8   &   5.6     &   24.9  \\  \hline\hline
Vanilla+Desc.                              &  7.0   &   2.5  &  1.2   &  0.7   &   5.2     &   12.0    &  9.4   &  3.9  &  1.8   &  0.7   &   4.6     &    18.9   \\  \hline
Desc. Aug.                                              &  10.8  &  4.6  &  2.0  & 1.1   &  6.0   &  21.6   &  10.0  &  4.3   &  2.0   &  1.1   &     4.9   &   21.3    \\  \hline
Web Aug.                                                   &  11.0   &  4.7   &  2.3   &  1.3   &   6.0     &   22.8    &  10.3   &  4.6   &  2.2   &   1.3  &   5.0     &  22.2    \\  \hline\hline
HL+Web Aug.                                                &   \textbf{11.7}  &  \textbf{5.1}   &  \textbf{2.6}   &  \textbf{1.6}   & \textbf{6.2}      &    \textbf{25.4}   &  \textbf{11.8}   &  \textbf{5.5}   &  \textbf{2.9}   &  \textbf{1.9}   & \textbf{5.7}  & \textbf{25.1}  \\  \hline
\end{tabular}
\label{table.VCperf}
\cuttabledown
\end{table*}

\normalsize
    	
\cutsubsectionup
\subsection{Results}

\noindent \textbf{Highlight sensitive captioner.}
When we apply our method on S2VT~\cite{SS15},
HL-1 significantly outperforms Vanilla and HL consistently improves over HL-1 (the better B@1-4, METEOR \hla{$6.2\%$}, and CIDEr \hla{$24.9\%$} in Table.~\ref{table.VCperf}).
When we apply our method on SA~\cite{VideoAttRNN}, the similar trend appears
and HL achieves the better METEOR \hla{$5.6\%$} and CIDEr \hla{$24.9\%$} than both of the Vanilla and the HL-1.
Moreover, the updated highlight detector \hla{(see technical report~\cite{VTGtech} for details)} achieves the best {$58.3\%$} mAP as compared to the initial {$54.2\%$} mAP.
We also found that training considering highlight temporal location is important, since Vanilla-GT-HL does not outperform Vanilla.
We further use the Vanilla model on S2VT to automatically select highlight clips. Then, we train a highlight-sensitive captioner based on these selected highlight clips as HL-0. It achieves METEOR \hla{$5.9\%$} and CIDEr \hla{$22.4\%$} which is only slightly inferior to HL on S2VT. It shows that our method trained without highlight supervision also outperforms Vanilla.

\noindent \textbf{Sentence augmentation.}
On VTW, when we apply our method on S2VT~\cite{SS15},
Vanilla+Desc. does not consistently improve accuracy; however,
both Web Aug. and Desc. Aug. improve accuracy significantly as compared to Vanilla (Table.~\ref{table.VCperf}).
When we apply our method on SA~\cite{VideoAttRNN}, the similar trend appears and Web Aug. achieves the best METEOR {$5\%$} and CIDEr {$22.2\%$}.
    	
    	\begin{figure}[!t]
	\cuttableup
	\centering
	\includegraphics[width=0.95\textwidth]{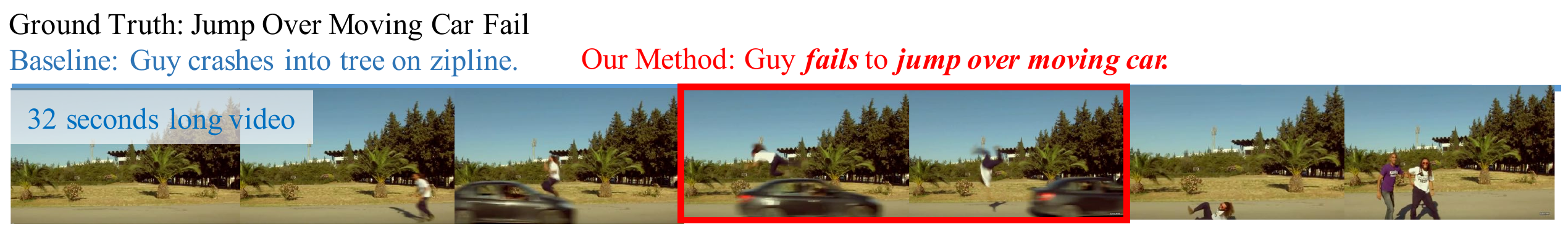}
	\includegraphics[width=0.95\textwidth]{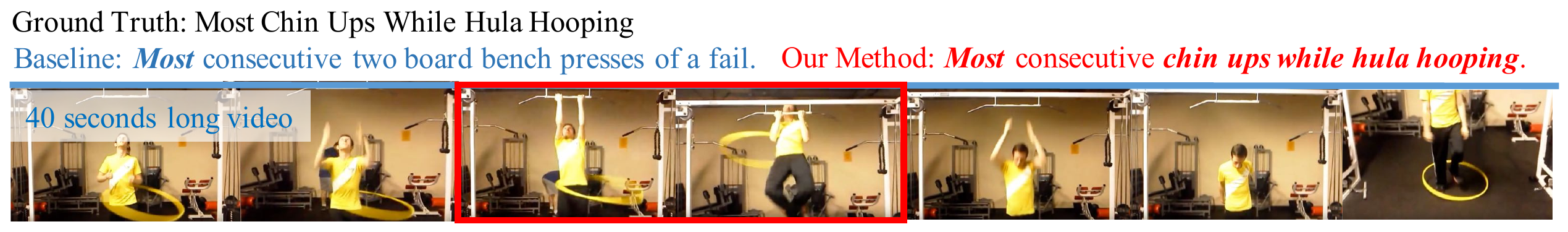}
         \includegraphics[width=0.95\textwidth]{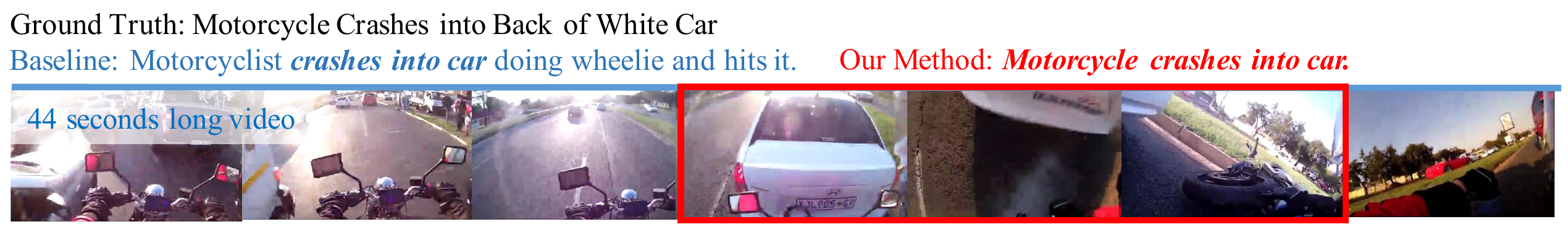}
    	\includegraphics[width=0.95\textwidth]{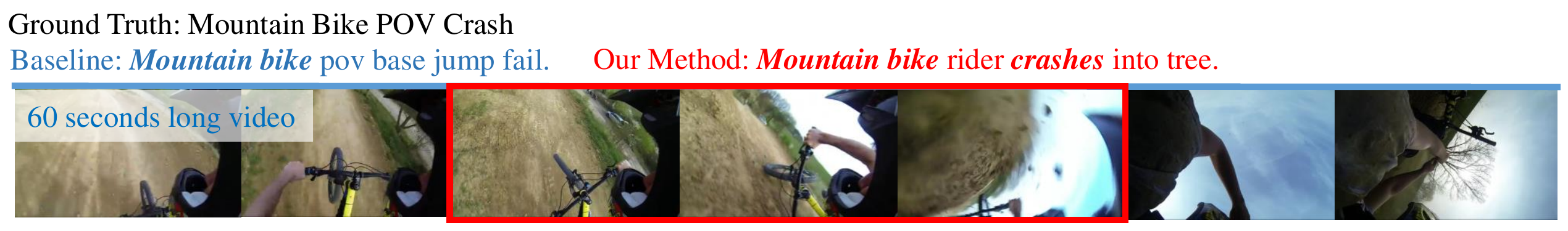}
    	\includegraphics[width=0.95\textwidth]{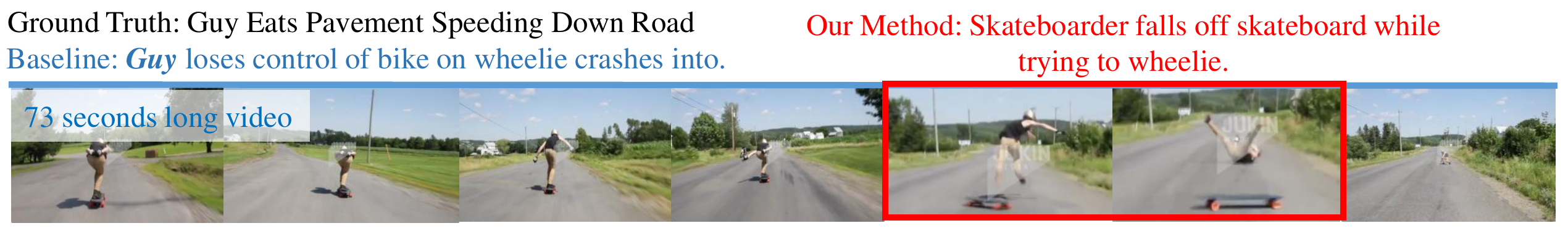}
	\includegraphics[width=0.95\textwidth]{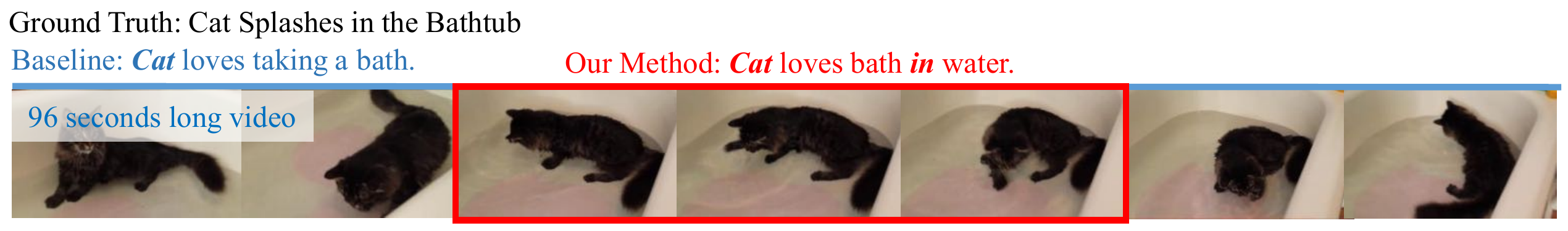}
    	\caption{\small Typical examples on VTW. Our method refers to ``HL+Web Aug. on S2VT". Baseline refers to ``Vanilla on S2VT". The words matched in the ground truth title are highlighted in bold and italic font. Each red box corresponds to the detected highlight with a fixed $3.3$ seconds duration.
    	Frames in the red box are manually selected from the detected highlight for illustration.
    	Note that our sentence in the last row has low METEOR, but was judged by human to be better than the baseline.
    	}\label{fig:ty}
	\vspace{-6mm}
    	\end{figure}

\noindent\textbf{Our full method.} On VTW dataset, HL with Web Aug. on both S2VT and SA outperform their own variants (last row in Table.~\ref{table.VCperf}),
especially in CIDER which gives higher value when a rare n-gram is predicted correctly.
Our best accuracy is achieved by combining HL with Web Aug. on S2VT.
We also ask human judges to compare sentences generated by our HL+Web Aug. on S2VT method and the S2VT baseline (Vanilla) on half of the testing videos \hla{(see technical report~\cite{VTGtech} for details)}.
\hla{Human judges decide that $59.5\%$} of our sentences are on par or better than the baseline sentences.
We show the detected highlights and generated video titles in Fig.~\ref{fig:ty}.
Note that our sentence in the last row of Fig.~\ref{fig:ty} has low METEOR, but was judged by human to be better than the baseline.

\noindent\textbf{Setence augmentation on M-VAD.}
Since S2VT outperforms SA in METEOR and CIDEr on VTW, we evaluate the performance of S2VT+Web Aug. on the M-VAD dataset~\cite{TorabiPLC15}.
Our method achieves $7.1\%$ in METEOR as compared to $6.6\%$ of the S2VT baseline and $6.7\%$ reported in \cite{SS15}. This shows its great potential to improve video captioning accuracy across different datasets.

\vspace{-4mm}
\section{Conclusion}
\vspace{-4mm}
We introduce video title generation, a much more challenging task than video captioning.
We propose to extend state-of-the-art video captioners for generating video titles.
To evaluate our methods, we harvest the large-scale ``Video Title in the Wild'' (VTW) dataset.
On VTW, our proposed methods consistently improve title prediction accuracy,
and the best performance is achieved by applying both methods.
\hla{Finally, on the M-VAD~\cite{TorabiPLC15}, our sentence augmentation method (METEOR  $7.1\%$) outperforms the S2VT baseline ($6.7\%$ in~\cite{SS15}).}

\small
\noindent\textbf{Acknowledgements.}
We thank Microsoft Research Asia, MOST 103-2218-E-007-025, MOST 104-3115-E-007-005, NOVATEK Fellowship, and Panasonic for their support. We also thank Shih-Han Chou, Heng Hsu, and I-Hsin Lee for their collaboration.
\normalsize


\end{document}